\documentclass[final,twocolumn]{elsarticle}

\usepackage{lineno,hyperref}
\modulolinenumbers[1]

\usepackage{booktabs} % For formal tables
\usepackage[linesnumbered,ruled,vlined]{algorithm2e}
\usepackage{multirow, makecell}
\usepackage{rotating}
\usepackage{amsmath}
\usepackage{subcaption}
\usepackage{adjustbox}
\usepackage{graphicx}
\usepackage{amssymb}
\usepackage{xcolor}
\usepackage[normalem]{ulem}

\newtheorem{definition}{Definition}

\SetCommentSty{mycommfont}
\journal{Journal of \LaTeX\ Templates}

\begin{document}

\begin{frontmatter}

 %\title{Elsevier \LaTeX\ template\tnoteref{mytitlenote}}
\title{Mining frequency-based sequential trajectory co-clusters}
% \tnotetext[mytitlenote]{Fully documented templates are available in the elsarticle package on \href{http://www.ctan.org/tex-archive/macros/latex/contrib/elsarticle}{CTAN}.}
% \tnotetext[mytitlenote]{The code and dataset are available in the github at https://.}
\tnotetext[mytitlenote]{The source code will be published if accepted for publication}

%% Authors
\author[mymainaddress,mythirdaddress]{Yuri Santos\corref{mycorrespondingauthor}}
\cortext[mycorrespondingauthor]{Corresponding author}
\ead{yuri.nassar@posgrad.ufsc.br}

\author[mysecondaryaddress,mythirdaddress]{Jônata Tyska}

\author[mymainaddress,mythirdaddress]{Vania Bogorny}

\address[mymainaddress]{Programa de Pós-Graduação em Ciência da Computação, Florianópolis}

\address[mysecondaryaddress]{Departamento de Informática e Estatística, Florianópolis}

\address[mythirdaddress]{Universidade Federal de Santa Cataina, Santa Catarina, Brazil}

% \address{Radarweg 29, Amsterdam}
% \fntext[myfootnote]{Since 1880.}

% \author[mymainaddress,mysecondaryaddress]{Elsevier Inc}

% \author[mysecondaryaddress]{Global Customer Service\corref{mycorrespondingauthor}}
% \cortext[mycorrespondingauthor]{Corresponding author}

% \address[mymainaddress]{1600 John F Kennedy Boulevard, Philadelphia}
% \address[mysecondaryaddress]{360 Park Avenue South, New York}

\begin{abstract}
Co-clustering is a specific type of clustering  that addresses the problem of finding groups of objects without necessarily considering all attributes. This technique has shown to have more consistent results in high-dimensional sparse data than traditional clustering. In trajectory co-clustering, the methods found in the literature have two main limitations: first, the space and time dimensions have to be constrained by user-defined thresholds; second, \emph{elements} (trajectory points) are clustered ignoring the trajectory sequence, assuming that the points are independent among them. To address the limitations above, we propose a new trajectory co-clustering method for mining semantic trajectory co-clusters. It simultaneously clusters the trajectories and their \emph{elements} taking into account account the order in which they appear. This new method uses the \emph{element} frequency to identify candidate co-clusters. Besides, it uses an objective cost function that automatically drives the co-clustering process, avoiding the need for constraining dimensions. We evaluate the proposed approach using real-world a publicly available dataset. The experimental results show that our proposal finds frequent and meaningful contiguous sequences revealing mobility patterns, thereby the most relevant \emph{elements}.
\end{abstract}

\begin{keyword}
Clustering analysis\sep Co-clustering\sep Trajectory analysis\sep Moving object trajectory
% \texttt{elsarticle.cls}\sep \LaTeX\sep Elsevier \sep template
% \MSC[2010] 00-01\sep  99-00
\end{keyword}

\end{frontmatter}

% \linenumbers

\section{Introduction}\label{sec:intro}
The recent advances on sensor's technology, the Internet of Things and mobile devices have contributed to the explosion in mobility data generation. Such a data can be collected in different ways such as GPS, wearables and sensor networks, social networks, which enables trajectory recording of humans, vehicles, animals, etc~\cite{Ferrero2016multiple,Beber2017individual}. 
The capability of analysing spatio-temporal sequences, also called moving object trajectories, benefits a wide range of applications and services, including traffic planning~\cite{Mohamed2016co}, animal behaviour analysis~\cite{%Li2010mining,
Shaham2015co}, location-aware advertising~\cite{Guo2016influence},  social recommendations~\cite{Bao2015recommendations}, medical diagnosis~\cite{Bayat2021gps}, to name just a few works.

The analysis of these data is the subject of several data mining techniques as, for instance, classification~\citep{Silva2019survey}, clustering~\citep{Yuan2017review}, frequent-pattern mining~\cite{Wang2020big}. A common challenge when dealing with mobility data is to find useful patterns while properly considering the sequence information~\citep{%Zheng2015overview,
Fournier2017survey,%Yang2018mining,
Bermingham2020mining}. Cluster analysis is one of the main categories of data mining tasks. It is a widely used technique in exploratory data analysis for outlier detection,  summarization, segmentation, etc~\cite{Saxena2017review}. Traditional clustering strategies are well known in the literature for producing a global analysis and considering all attributes in the clustering process. An example is the well-known algorithm \textit{k-means}~\citep{Wagstaff2001}. Over the years, with the increase of data volume and heterogeneity, employing clustering methods has become more challenging since a high number of attributes and sequential data can increase the computational complexity and affect the cluster results consistency. This well-known problem is called \textit{curse of dimensionality}~\cite{Xu2005survey}.

To overcome this challenge, co-clustering approaches can be used, as they simultaneously cluster rows and columns of a matrix~\citep{Salah2019directional} without exploring all the possible row-columns combinations. Co-clustering deals with the curse of dimensionality problem by clustering the main data dimension, the objects (rows), while applying dimensionality reduction techniques on the other dimension, the features (columns). The focus relies on finding subsets of objects and attributes that can represent submatrices (co-clusters)~\citep{%Hartigan1972direct,Govaert2003clustering,
Madeira2004biclustering}. In order to accomplish, efficient strategies are needed, since discovering all significant co-clusters in a given matrix is proved as an NP-hard challenging problem~\citep{Cheng2000biclustering}. 

Different trajectory co-clustering approaches were proposed to deal with trajectory data regarding the spatial dimension~\cite{Sankaranarayanan2010learning,Shaham2015co,Mohamed2016co}, space and time ~\cite{Han2017linking}, or \emph{semantics}~\cite{Arian2017characterizing,Wang2017vessel,Hu2019nonnegative}. However, these approaches have two main limitations. First, the space and time dimensions are constrained using thresholds, which limits the method's capacity to find patterns. For instance, thresholds or grids are used to assess the distance/similarity between spatial locations as well as using time windows or time-slices to divide the temporal dimension. Second, \textit{elements} (trajectory points) are clustered ignoring the trajectory sequence, assuming that the points are independent among them. This ignores the fact that the order in which the \emph{elements} appear on a trajectory have an important meaning. 
Noting this gap in the literature, we propose a new trajectory co-clustering method called SS-OCoClus for mining semantic trajectory co-clusters dealing with these two limitations. The new method uses the \textit{elements} frequency to identify candidate co-clusters, and then checks the \emph{element} order in these candidates. Besides, it uses an objective cost function that automatically drives the co-clustering process, avoiding the need for thresholds that constrain dimensions.

We hypothesize that the order of visited \textit{elements} in a sequence, together with its overall frequency are important to reveal mobility patterns and hidden user behaviors. In order to make this type of co-clustering analysis computational feasible, we use the insight of combining overall frequency and the \emph{elements} order for efficiently finding order-aware co-clusters. Sequences composed by low frequency \emph{elements} are not considered seeking computational efficiency. We assume that sequences formed by high-frequency \emph{elements} are more meaningful, since they often represent the preference for visiting specific places, or relevant trajectory dynamics repeated over time. Therefore, we suppose that analyzing only the most visited \emph{elements} will provide a higher probability of identifying order-aware co-clusters that represent relevant patterns. We evaluate SS-OCoClus on a dataset from the Foursquare social network~\cite{Yang2014FS-NY} %,Petry2019towards,Hu2019nonnegative} 
to demonstrate its validity and efficacy. Once the trajectory co-clustering methods in the literature do not take into account the \emph{elements} order, we show the number of relevant co-clusters, grouped \emph{elements} and the overall entropy of our clustering results. In summary, we make the following contributions:

\begin{itemize}
    \item Address the current trajectory co-clustering limitation that does not consider the order of the \emph{elements} in trajectories. So, we propose an incremental trajectory co-clustering method to find trajectory co-clusters giving their \textit{most similar points}. Instead of dealing with sparse and highly-dimensional matrix, our approach incrementally identifies frequent sequences to form candidate co-clusters;
    
    \item Make use of both the \emph{elements} order and their frequency information to discover useful mobility patterns that are visualized with the alluvial diagram;
    
    \item Identify similar mobility patterns over semantic locations, enabling cluster objects with the same behaviour even when they pass in different areas and move at different times.
    
\end{itemize}

The remainder of this work is organized as follows. The basic concepts and related works are presented in Section~\ref{sec:bg_ps}. Section~\ref{sec:proposal} presents the details of our method. Section~\ref{sec:experiment} presents the clustering results of our proposal. Finally, the conclusion and further research directions of our work are presented in Section~\ref{sec:conclusion}.

\section{Basic Concepts and Related Works}\label{sec:bg_ps}
In this section we present the basic concepts to guide the reader throughout this paper. We start by providing basic definitions for \textit{semantic trajectory} , defined according to \citep{Furtado2016multidimensional}, co-clustering, and a comprehensive understanding on the state-of-art related methods. 

\subsection{Basic Concepts}

\begin{definition}{\textbf{Semantic Trajectory} ST:} A semantic trajectory ST is a time ordered sequence of elements $ST = \langle e_{1},e_{2},\ldots e_{n}\rangle$, where each element $e_{i}$ has a set of attributes \{$d_{1}$,$d_{2}$,\ldots,$d_{l}$\} characterizing it according to $l$-dimensions.
\end{definition}

Semantics is any type of information associated to mobility data other than spatial location and time. Several semantic attributes can be added to the \emph{elements} such as the activity, the category of the POI, the weather condition, etc. A dataset with $N$ sequences can be represented as a matrix $D$ with $N$ rows and $M$ columns, where each entry of $D$ represents an \textit{element} of a sequence. An \textit{element} $t_{ij}$ of $D$ ($1\leq i \leq N$ and $1\leq j \leq M$) is equal to the number of times that the $j$-th \textit{element} occurred in the $i$-th sequence. Co-clustering is the grouping task of finding $K$ co-clusters in $D$ where each co-cluster is formed by a subset of rows and columns~\citep{Madeira2004biclustering}. %,Govaert2013CoclusModels}. 
A subset of rows $I$ can be represented as a binary vector of length $N$, where $I_{i} = 1$ indicates that the $i$-th row is present in $I$. Similar to that, a subset of columns $J$ with $J_{j} = 1$ indicates that the $j$-th column is present in $J$. More formally, a co-cluster can be defined as follows:

\begin{definition}{\textbf{Co-cluster} C:}\label{def:co-cluster} Let $D$ be a matrix, $I$ be the subset of rows, $J$ be the subset of columns; a co-cluster $C$ is defined as $C = \langle I,J \rangle$. The entries $c_{ij}$ of co-cluster $C$ are formed by the outer product of its subsets $I$ and $J$. Thus, a co-cluster $C$ can represent a submatrix of $D$.
\end{definition}

We also make use of the overlap coefficient~\cite{Vijaymeena2016survey}. Formally, it is defined as follows:

\begin{definition}{\textbf{Overlap Coefficient} Oc:}\label{eq:over_coef}
Given two sets $A$ and $B$, the overlap coefficient is defined as the size of intersection of $A$ and $B$ over the size of the smaller set between $A$ and $B$.
\end{definition}

\subsection{Related Works}

% Space:
Sankaranarayanan and Davis~\cite{Sankaranarayanan2010learning} proposed a mutual information co-clustering method that simultaneously clusters the start and end locations of the pedestrian trajectories. This approach is limited to the paired analysis once it clusters the start and end locations with similar probability in the matrix. Shahan~\cite{Shaham2015co} proposed a co-clustering method to find co-clusters with a fuzzy strategy that groups objects with a lagged (delay) pattern. This approach cannot deal with semantic data and cluster the attributes regardless of the visited order, thereby overlooking hidden mobility patterns. Mohamed et al.~\cite{Mohamed2016co} used a co-clustering method to group trajectories and road segments in the context of the road network. However, this approach focus on identifying similar trajectories with common road segments regardless of the visited order in the sequence.

% Spatio-temporal:
Han et al.~\cite{Han2017linking} proposed a framework designed for criminal investigations by employing a spectral co-clustering approach on access trajectories in social networks. This approach does not group the trajectories, and it uses the spatial and temporal dimensions independently to identify groups of user IDs. Besides that, it does not deal with semantics, the visited order in the trajectories, and constrains the temporal dimension using a time window.

% Semantic:
Arian et al.~\cite{Arian2017characterizing} used a co-clustering method to group users and activities from the origin-destination location of moving objects. Once it groups users instead of trajectories, the approach cannot find movement patterns in the trajectories. Besides, it groups the \emph{elements} regardless of the visited order in the trajectories, therefore, misinterpreting the mobility patterns. Wang et al.~\cite{Wang2017vessel} used a sparse bilinear decomposition and a sparse multi-linear decomposition to find co-clusters in vessel trajectories. The multi-linear decomposition using tensor groups the regions, time-slices, and ship type in the set of trajectories. Besides, this method does not consider the sequence in the trajectories, and it constrains the temporal dimension. Hu et al.~\cite{Hu2019nonnegative} proposed a co-clustering method based on Nonnegative Matrix Tri-factorization for grouping users and POIs. This method focuses on grouping the POIs regardless of the visited order in the trajectory. Besides, it constrains the time dimension, which misinterprets the movement patterns.

\section{Trajectory Co-clustering Approach}\label{sec:proposal}
In this section we present SS-OCoClus, a order-aware frequency based co-clustering method for semantic trajectories. SS-OCoClus finds candidate co-clusters based on frequent sequential \emph{elements} that are then evaluated to keep the most relevant candidates as the semantic co-clusters. Furthermore, the proposed method uses a cost function and an overlap threshold to evaluate the candidate co-clusters generated for each frequent sequence found in the dataset. In the following we present the method definitions in section \ref{sec:met_definition} and the method details in section \ref{sec:met_description}.

\subsection{Main definitions}\label{sec:met_definition}

SS-OCoClus forms candidate co-clusters by testing each \emph{element} incrementally. It starts with the most frequent \emph{elements} and then expands the sequence size by adding one frequent \emph{element} at a time. More formally, a candidate co-cluster is defined as follows:

\begin{definition}{\textbf{Candidate Co-cluster} CC:}\label{def:cand_seq}
Let $seq$ be a sequence of \textit{elements}, $EM$ be the elements mapping inverted index, and $TM$ be the trajectories mapping inverted index, a candidate co-cluster $CC$ is a tuple $CC = \langle S_{TM}, S_{seq}\rangle$, where $S_{seq}$ is a sequence of \textit{elements} that exist in $TM$ with intersection in $EM$, and $S_{TM}$ is a subset of trajectories indices that contains the sequence $S_{seq}$.
\end{definition}

SS-OCoClus evaluates each candidate co-cluster to keep only the most relevant ones as the semantic trajectory co-clusters. The candidates relevance can be defined in terms of the number of trajectories, cost value, or both as the reference. The measured relevance is compared with a given statistical metric, that can be the average or the $z$-score. More formally, the semantic co-cluster is defined as follows:

\begin{definition}{\textbf{Semantic Co-cluster} SC:}\label{def:semantic_co-cluster}
An order-aware semantic trajectory co-cluster SC is a candidate co-cluster CC with the relevance not smaller than the used statistical metric. Thus, the semantic co-cluster SC is a subset of trajectories (objects) and \textit{elements} (attributes), where the \textit{elements} represent a frequent contiguous (sub) sequence in the dataset.
\end{definition}

Supposing that there will be several candidate co-clusters in a dataset, it is not a trivial task to manually define thresholds, such number of rows or columns, to generate candidate co-clusters. Besides, the optimality of a candidate can be expressed as an objective function $\mathcal{F}(I,J)$ that can rank it. Such common functions when performing co-clustering use the logic of perimeter $\mathcal{F}(I,J) = |I|+|J|$ or area $\mathcal{F}(I,J) = |I|\times|J|$~\cite{Avraham2004CostFunc}. Similarly, Lucchese et al.\cite{Lucchese2013unifying} proposed a cost function that combines the perimeter and area into one general function to mine frequent patterns binary datasets.

Inspired by the work of Lucchese et al.\cite{Lucchese2013unifying}, we adapted their cost function for tackling order-aware co-clustering. We consider the number of overlapped \emph{elements} instead of counting the number of \textit{false positives} (matrix entries equal to 0 present in a pattern) and \textit{false negatives} (matrix entries equal to 1 not present in a pattern) as noise data. More formally, the number of overlapped \emph{elements} $Cov$ is defined as follows:

\begin{definition}{\textbf{Overlapped Elements} $Cov$:}\label{def:over-elem-cov}
Given the set of candidate co-clusters $\Phi$ and the candidate co-cluster CC, the number of overlapped elements $Cov$ is defined as the number of intersected elements between CC and $\Phi$.
\end{definition}

SS-OCoClus uses the cost function in order to evaluate the candidate co-cluster generated by the frequent sequence. Thus, we define the new cost function $\mathcal{F}$ as follows:

\begin{definition}{\textbf{Cost Function} $\mathcal{F}$:}\label{def:cost_func_f}
 Let $\Phi$ be a set of co-clusters, $CC$ be a candidate co-cluster, $|CC_{I}|$ be the size of a subset of trajectories, $|CC_{J}|$ be the length of the co-cluster sequence, and $Cov$ be the number of overlap elements between $CC$ and $\Phi$; 
a cost function $\mathcal{F}$ is defined as follows:
\begin{equation}\label{eq:cost_func}
    \mathcal{F}(CC,\Phi) = (|CC_{I}|+|CC_{J}|) - (|CC_{I}|\times|CC_{J}|) + Cov
\end{equation}
\end{definition}

\subsection{The proposed Method}\label{sec:met_description}

\autoref{alg:tracoclus} shows the organization of our trajectory co-clustering method. It receives three inputs: trajectory dataset $D$, maximum number of candidate co-clusters $K$, an overlap threshold $\epsilon$ between co-clusters, statistical metric $stat\_met$, and relevance reference $rel\_ref$. As output, it returns a set of co-clusters $\Phi$. It starts by initializing the set of co-clusters $\Phi$ (line 1), then the \textit{elements} mapping ($EM$), trajectory mapping ($TM$) and \emph{elements} frequency ($Els\_freq$) data structures (line 2) by preprocessing $D$. Remember that $EM$ and $TM$ are inverted indexes, e.g., dictionaries for simplicity.
The $EM$ structure stores in each $element_{m}$ the trajectories ID that contain the $element_{m}$, while $TM$ stores the \emph{elements} order for each trajectory, and $Els\_freq$ stores the overall frequency for each $element_{m}$ in the dataset $D$. These data structures form the basis for our method to identify frequent contiguous sequences in $D$. The algorithm iterates at most $K$ times (line 3), as this parameter determines the maximum number of candidate co-clusters to be extracted. Rather than testing all the possible $2^{M}$ combinations of \textit{elements}, these are sorted by their frequency in descending order (from the most frequent to least) to maximize the probability of identifying frequent sequences in the clustering process (line 4). 
So, we avoid to test all combinations by focusing on the high frequencies. 
Next, the subset of trajectories and \emph{elements} are initialized as empty (line 5) to form the candidate co-clusters $CC$ and $CC^{*}$ (line 6). These candidates are used to store the identified frequent sequence pattern.

\begin{algorithm}[!ht]
\DontPrintSemicolon
\SetNoFillComment
\footnotesize

\SetKwInOut{Input}{Input}
\SetKwInOut{Output}{Output}
\Input{Trajectory dataset $D$, Max number of candidate co-cluster $K$, Overlap threshold $\epsilon$, Statistical metric $stat\_met$, and Relevance reference $rel\_ref$}
\Output{Set of co-clusters $\Phi$}

$\Phi \leftarrow \{\emptyset\}$\;
$EM,TM,Els\_freq \leftarrow initializeData(D)$\;
\For{candidate\_iter = 0 to K}{
    $sort(Els\_freq,desc)$ \tcp*{descending order}
    $I,J \leftarrow [\emptyset]$ \tcp*{subset of trajectories and \textit{elements}}
    $CC^{*} \leftarrow CC,CC \leftarrow \langle I,J \rangle$ \tcp*{co-clusters}
    $sequence\_cc \leftarrow [\emptyset]$\tcp*{cluster  sequence}
    $Els\_queue \leftarrow queue(Els\_freq)$\;
    \tcc{step to find a candidate}
    \For{i = 0 to Els\_queue.length}{
        $el\_p \leftarrow Els\_queue.pop()$\;
        $sequence\_cc.append(el\_p)$\;
        $Els\_queue.append(el\_p)$\;
        $Els\_to\_test \leftarrow Els\_queue.length$\;
        \tcc{step to expand candidate sequence}
        \While{Els\_to\_test \textgreater 0}{
            $el\_q \leftarrow Els\_queue.pop()$\;
            $Els\_queue.append(el\_q)$\;
            $CC^{*} \leftarrow candidateCC(sequence\_cc,el\_q,EM,TM)$\;
            \eIf{$\mathcal{F}$($CC^{*},\Phi$) $\leq$ $\mathcal{F}$($CC,\Phi$) $\And max(Oc(CC^{*},\Phi)) \leq \epsilon$}{
               $CC \leftarrow CC^{*}$\;
               $Els\_to\_test \leftarrow Els\_queue.length$\;
               $sequence\_cc.append(el\_q)$\;
            }{
               $Els\_to\_test$ -= 1\;
            }
        }
        \eIf{$CC == \emptyset$}{
            $sequence\_cc.pop()$\tcp*{test the next el\_p}
        }{
            \textbf{break}\tcp*{candidate identified}
        }
    }
    \eIf{$\mathcal{F}(CC,\Phi) \geq 0$}{
        \tcc{keep the most relevant candidates}
        $prune(\Phi,stat\_met,rel\_ref)$\;
        \textbf{break}\tcp*{No relevant candidates anymore}
    }{
        $\Phi.append(CC)$\tcp*{store the candidate}
        $Els\_freq.update(CC)$\tcp*{update frequencies}
    }
}

\KwRet{$\Phi$}
 \caption{order-aware co-clustering}\label{alg:tracoclus}
\end{algorithm}

SS-OCoClus uses the \textit{element} frequency to identify frequent sequences, assuming that for a sequence to be frequent, the \emph{elements} that belong to the sequence will also be frequent as \textit{a priori} principle~\cite{Bastide2000mining}. The method creates a queue $Els\_queue$ using the \textit{elements} frequency (line 8) aiming to expand the candidate co-cluster sequence. SS-OCoClus starts adding the \emph{element} $el\_p$ in the co-cluster sequence (line 10) to test it with the next \emph{element} in the queue ($el\_q$). The method tries to expand the frequent sequence while the \emph{elements} queue does not reach its end, when there is no \emph{element} to test(line 14). Then, the element $el\_q$ is dequeued (line 15) and it is added to the end of the queue for further analysis (line 16). SS-OCoClus uses the $candidateCC$ function to identify frequent sequences by expanding the candidate co-cluster sequence $sequence\_cc$ with the element $el\_q$ regarding the intersection of each \emph{element} in $EM$ and the order-aware of each trajectory in $TM$ (line 17). If the frequent sequence exits, its pattern is used as a candidate co-cluster.

The $candidateCC$ function tries to form two frequent sequences: (i) one sequence in the form of $sequence\_cc \rightarrow el\_q$, and (ii) another sequence in the form of $el\_q \rightarrow sequence\_cc$ (we use $\rightarrow$ to mean \textit{from \textbf{A} to \textbf{B}}). These sequences can represent a pattern formed by a subset of trajectories and \textit{elements} of $D$. From that, given the $sequence\_cc$ and $el\_q$ that form the sequence of elements, the \emph{elements} map $EM$, and the trajectories map $TM$, the $candidateCC$ function returns the frequent sequence pattern with the smallest cost value representing the candidate co-cluster. Next, it checks if the candidate co-cluster $CC^{*}$ cost value is smaller than the candidate co-cluster $CC$ and if its maximum overlap coefficient $Oc$ does not exceed the overlap threshold $\epsilon$ (line 18). If this condition is respected the method performs as follows: the candidate co-cluster is accepted, and SS-OCoClus updates $CC$ by replacing it with $CC^{*}$ (line 19); reassign the counter number with the queue length (line 20); update the candidate co-cluster sequence (line 21). Otherwise, it decrements the counter number of \emph{elements} to test (line 23).

After checking all \emph{elements} in the queue trying to expand the candidate co-cluster sequence (line 14), if the candidate $CC$ is empty (line 24) the method removes the \emph{element} $el\_p$ (line 25) and repeats the loop (line 9) to test the next $el\_p$ \emph{element} in the sequence. Otherwise, it stops searching for new candidates (line 27). Next, the method checks if the candidate $CC$ does not have the minimum cost value to accept it as a candidate co-cluster (line 28). The method identifies it automatically thanks to the cost function $\mathcal{F}$ stated in \autoref{def:cost_func_f}.

Considering the candidate $CC$, if it does not obtain a minimum cost value (line 28), SS-OCoClus \textit{prunes} the set of candidates to keep the most relevant candidates as the semantic co-clusters $SC$ in $\Phi$ (line 29). For that, SS-OCoClus allows the user to specify the relevance reference $rel\_ref$, where it can consider the number of trajectories, cost value, or both. Besides, the user can choose between two statistical metrics $stat\_met$ to prune the candidate result: first, in terms of the average; and second, in terms of the number of deviations ($z$-score). After this, the method returns the most relevant semantic co-clusters (line 34) regarding the relevance reference $rel\_ref$ and the statistical metric $stat\_met$. However, if candidate $CC$ has a minimum cost value, it is appended to the set of candidates co-clusters $\Phi$ (line 32), and the $Els\_freq$ is updated regarding the $CC$ (line 33). SS-OCoClus updates $Els\_freq$ decrementing the \emph{element} frequency in the $Els\_freq$ by counting the times that an \emph{element} appears in the co-cluster sequence and multiplies by the number of trajectories in the candidate. Thus, the process for finding candidate co-clusters is repeated at most $K$ times, but it also stops when the one last analysed candidate does not reduce the cost function at a minimum value.

\section{Experimental Evaluation}\label{sec:experiment}
In this section we evaluate the proposed method over a real trajectory dataset with characteristics aiming to demonstrate the efficacy and usefulness of SS-OCoClus on finding order-aware co-clusters. We use three $z$-score values (-1, 0, and 1) as the minimum threshold to keep the most relevant candidates. Besides, using $z = 1$ keeps candidates with high number of trajectories and small cost value reducing the number of cluster. Thus, due to space limitation, we focus on presenting the clustering results considering $z = 1$. The experiments are performed in a machine with a processor Intel i7-7700 3.6GHz, 16 GB of memory, and operational system Windows 10 64bits.

\subsection{Datasets}

\begin{table}[ht]
    \centering
        \caption{Datasets description (averages reported in the format avg$\pm$std).}\label{tab:datasets_ss-ococlus}
        \resizebox{.475\textwidth}{!}{%\begin{minipage}{\textwidth}
        \begin{tabular}{l|c|c|c|c|c}
        \hline
        \textbf{Dataset} & \scalebox{.85}{\textbf{\makecell{Num. of\\trajectories}}} & \scalebox{.85}{\textbf{\makecell{Num. of\\unique Elements}}} &  \scalebox{.85}{\textbf{\makecell{Num. of \\users}}} &
        \scalebox{.85}{\textbf{\makecell{Avg. \# of traj.\\per user}}} & \scalebox{.85}{\textbf{\makecell{Avg. traj.\\length per user}}} \\
        \hline
        FS-NY$_{top193}$ & 3079 & 491 & 193 & 15.95$\pm$6.33 & 20.52$\pm$8.06 \\
        \hline
        FS-NY$_{top81}$ & 1749 & 437 & 81 & 21.59$\pm$6.09 & 22.81$\pm$9.66 \\
        \hline
        FS-NY$_{top10}$ & 352 & 310 & 10 & 35.2$\pm$4.4 & 32.23$\pm$15.46 \\
        \hline
        % \bottomrule
        \end{tabular}
        % \end{minipage}
        }
\end{table}

\begin{table*}[ht]
    \centering
        \caption{Co-clustering result for each dataset over different candidate reference using $z = 1$. Average and coefficient of variation are in the form AVG[CV\%] and we assume the absolute value of CV.}
        \label{tab:exp:main_results}
        \resizebox{.8\textwidth}{!}{%\begin{minipage}{\textwidth}%.81\textwidth
        \begin{tabular}{cccccccccc}
        \toprule
        \multirow{1}{*}{Dataset} & \makecell{Relevance\\reference} & \makecell{Number of\\co-clusters} & \makecell{AVG number\\of trajectories} & \makecell{AVG cost\\} & \makecell{Number of\\elements} &
        \makecell{Sequence\\length} & \makecell{AVG number\\of users} & \makecell{Overall\\entropy} \\
        \hline
        %%% all users %%%
        \multirow{3}{*}{\rotatebox[origin=c]{0}{FS-NY$_{top193}$}}
        %%%%%%%%% Z= 1
        & Trajectory & 69 & 67.46 [60] & -41.3 [102.1] & 32 & 2.09 [13.39] & 24.71 [42.65] & 6.31 \\
        & Cost$^\blacktriangle$ & 49 & 75.04 [61.55] & -55.37 [78.49] & 28 & 2.29 [26.63] & 25.65 [48.49] & 5.06 \\
        & Both & 77 & 62.88 [64.67] & -40.3 [99.42] & 36 & 2.21 [23.52] & 22.96 [49.65] & 6.45 \\
        \hline
        \hline
        %%% top 81 %%%
        \multirow{3}{*}{\rotatebox[origin=c]{0}{FS-NY$_{top81}$}}
        %%%%%%%%%% Z = 1
        & Trajectory & 53 & 54.08 [60.5] & -34.53 [99.32] & 25 & 2.11 [14.99] & 16.68 [39.44] & 5.89 \\
        & Cost$^\blacktriangle$ & 34 & 61.97 [63.88] & -47.97 [77.11] & 34 & 2.32 [29] & 15.35 [49.58] & 4.16 \\
        & Both & 57 & 52.21 [63.54] & -34.35 [96.48] & 29 & 2.21 [25.05] & 15.77 [45.68] & 5.94 \\
        \hline
        \hline
        %%% Top 10 %%%
        %%%%%%%%% Z = 1
        \multirow{3}{*}{\rotatebox[origin=c]{0}{FS-NY$_{top10}$}}
        & Trajectory & 22 & 22.59 [34.83] & -19.68 [50.15] & 21 & 2.5 [31.2] & 2.18 [47.24] & 1.02 \\
        & Cost$^\blacktriangle$ & 27 & 18.22 [53.01] & -21.89 [39.74] & 26 & 3.22 [34.16] & 1.70 [61.76] & 0.83 \\
        & Both & 33 & 18.39 [48.12] & -19.42 [49.07] & 28 & 3 [36.66] & 1.79 [56.42] & 1.02 \\
        \bottomrule
        \multicolumn{7}{l}{Cost$^\blacktriangle$: It uses the other corresponding side of $z$-score, i.e., -1 when $z$=1 and vice-versa.}
        \end{tabular}
        % \end{minipage}
        }
\end{table*}

SS-CoClus is evaluated over three datasets extracted from the Foursquare NY dataset~\cite{Yang2014FS-NY}, as detailed in~\autoref{tab:datasets_ss-ococlus}. It shows the total number of trajectories, the number of unique \emph{element} values, the number of users, the average number and length of trajectories per user. This dataset contains check-ins of users collected from April 2008 to October 2010. Furthermore, it is composed by weekly trajectories of check-ins for each user, given the whole set of check-ins.%~\cite{Petry2019towards}. 
The Foursquare API\footnote{https://developer.foursquare.com/} provided the semantic information related to the POI: category (\textit{root-type}), subcategory (\textit{type}), place name (\textit{poi}), and \textit{price} (a numeric classification); the Weather Wunderground API\footnote{https://www.wunderground.com/weather/api/} provided the \textit{weather} conditions. We select the POI subcategory (\textit{type}) to generate the trajectory sequences for all three datasets which contains 491 unique \emph{elements}. It is an intermediate dimension of semantic POI information, i.e., it is not too specific as the place name dimension and it is not too general as the category dimension.

\subsection{Result and Analysis}

We consider that an \emph{element} is frequent if its frequency is equal or higher than the average. Therefore, the number of unique \emph{elements} considered as frequent are, respectively, 58, 88, and 93 for the datasets FS-NY$_{top10}$, FS-NY$_{top81}$, and FS-NY$_{top193}$.~\autoref{tab:exp:main_results} presents the co-clustering results considering seven characteristics: (i) the relevance reference; (ii) the number of final co-clusters regarding the candidate reference; (iii) the average number of trajectories; (iv) the average cost value; (v) the number of unique clustered \emph{elements}; (vi) the average sequence length; (vii) the average number of unique users in the co-clusters; and (x) the overall entropy of the set of co-clusters. The number of trajectories and the cost value are inversely proportional. So, we combine both using the other corresponding side of the $z$-curve when the reference is the cost value. For example, if $z$ equals 1, it is set to -1 for the cost value reference and vice-versa. Thus,~\autoref{tab:exp:main_results} shows the trajectory co-clustering results over different candidate reference using $z = 1$.

In FS-NY$_{top193}$, the number of co-clusters are 69, 49, and 77 for trajectory, cost, and both, respectively. For FS-NY$_{top81}$, the number of co-clusters are 53, 34, and 57 for trajectory, cost, and combine, respectively. In FS-NY$_{top10}$, the number of co-clusters are 22, 27, and 33 for trajectory, cost, and both, respectively. Using the cost as reference identifies the smallest number of co-clusters in FS-NY$_{top193}$ and FS-NY$_{top81}$. It means that these co-clusters have the highest balance between the number of trajectories and the sequence length. Regarding the clustered \emph{elements}, we may notice that less than 50\% of the \emph{elements} are relevant to identify frequent sequences and forming semantic co-clusters. For example, SS-OCoClus uses 93 \emph{elements} to identify the co-cluster in FS-NY$_{193}$. However, the maximum number of unique \emph{elements} in the final result was 36 when pruning the candidates using both trajectories and the cost as reference. Therefore, 36 \emph{elements} are relevant and contributes to identify 77 semantic co-clusters. Furthermore, in FS-NY$_{top81}$ dataset, considering the trajectory, cost and combine references, the number of unique clustered \emph{elements} are 25, 34, and 29, respectively. In the same way, for the FS-NY$_{top10}$ dataset, the number of unique  clustered \emph{elements} are 21, 26, and 28, respectively.

It can be seen in \autoref{tab:exp:main_results} that the majority of the average sequence length values are less than 3 \emph{elements}, and only in the FS-NY$_{top10}$ dataset was identified an average equal or greater than 3. Using the cost as the co-cluster reference leads to the discovering of co-clusters with long sequences without discarding many trajectories. Besides, in this dataset, we may infer that it is not common to have long frequent mobility patterns. In addition, the coefficient of variation of the sequence length between co-clusters considering all datasets spans from 13.39\% up to 36.66\%. Such variation within the clustering shows that the co-clusters are heterogeneous regarding the sequence length. Another observation is the high heterogeneity between co-clusters regarding the average number of unique users per cluster. It spans from 39.44\% to 61.76\%, where the minimum value occurs in FS-NY$_{top81}$ considering 34 clusters, while the maximum value occurs in FS-NY$_{top10}$ with 27 clusters.

\section{Conclusion}\label{sec:conclusion}
We proposed SS-OCoClus, a new trajectory co-clustering method that finds frequent sequences regarding the order and frequency of the \emph{elements}. The main contribution of the new method is that it groups trajectories containing a sequence of similar \emph{elements} that appear in the same order and that are frequent along different trajectories. This is a novelty compared to available trajectory co-clustering methods  that only find groups of trajectories that contain similar \emph{elements}. The method is  evaluated and validated on a real-world semantic trajectory dataset. Furthermore, despite these initial results were limited to analyzing only one semantic dimension, the new method can deal simultaneously with multiple aspects of  trajectories by concatenating different dimensions. Moreover, SS-OCoClus is driven by a cost function that allows identifying the number of co-clusters automatically. The experiments and results quantitatively and qualitatively showed the efficacy and utility of the proposed method.

It is noteworthy that clustering analysis, a common descriptive data mining task, is often exploratory in nature\cite[chapter 1]{tan2016introduction}. So, based on the experimental results, three main insights could be discovered: (i) on average, the trajectory patterns do not have long sequences; (ii) SS-OCoClus shows that a small number of \emph{elements} is relevant to identify sequences; and (iii) using the frequency of the \emph{elements} allows finding frequent sequences revealing different mobility patterns. We argue that these results are interesting because they reveal behaviours that existing co-clustering methods in the literature are unable to find.

After obtaining these first encouraging results, we envisage some future directions of investigation. First, the final number of co-clusters depends on the number of identified candidate co-clusters $K$, used by the cost function to drive the clustering process. Instead of requiring it to be manually specified by the user, we intend to test some statistical strategy for automatically inferring the useful number $K$. Second, SS-OCoClus deals with any type of sequential data, but when analyzing multiple aspects trajectories, for dealing with multiple dimensions, it requires manually-defined dimension concatenation. As another future work we will  expand SS-OCoClus to automatically find relevant dimensions in multiple aspect trajectories. Third, SS-OCoClus prunes the set of candidates using statistical metrics and candidate reference. This means that by using different hyperparameters, different co-clustering results will be generated. Therefore, it might be interesting to automatically  test multiple statistical strategies to keep the most relevant co-clusters, generating useful results without requiring \textit{a priori} user knowledge.

\section*{Acknowledgments}\label{sec:tkxTo}
This work was financed  by the Brazilian agencies Coordenação de Aperfeiçoamento de Pessoal de Nivel Superior - CAPES (Finance code 001), Conselho Nacional de Desenvolvimento Científico e Tecnológico - CNPq, and supported by the MASTER project that received funding from European Commission’s Horizon 2020 research and innovation programme under the Marie Sklodowska-Curie grant agreement N. 777695.

% \bibliography{mybibfile}
\bibliography{myRef}

\end{document}